\documentclass{article}

\PassOptionsToPackage{numbers}{natbib}
 \usepackage[preprint]{neurips_2026}

\usepackage[utf8]{inputenc} 
\usepackage[T1]{fontenc}    
\usepackage{hyperref}       
\usepackage{url}            
\usepackage{booktabs}       
\usepackage{amsfonts}       
\usepackage{nicefrac}       
\usepackage{microtype}      
\usepackage{xcolor}         
\usepackage{graphicx}
\usepackage{multirow}
\usepackage{makecell}
\usepackage{amsmath}
\usepackage{enumitem}
\usepackage{wrapfig}
\usepackage[table]{xcolor}

\definecolor{incgray}{gray}{0.55}
\definecolor{bestgreen}{RGB}{217,242,217} 
\definecolor{darkedred}{RGB}{146,57,49}
\definecolor{Forestgreen}{RGB}{34,139,34}

\usepackage{subcaption}

\title{RAO-Nav: Probing Omni-Language Models for Zero-shot Semantic Audio-Visual Navigation}

\author{%
  Qilang Ye\textsuperscript{1,2},
 Meng Liu\textsuperscript{3,2 *},
  Yu Zhou\textsuperscript{1,2 *}
  \\[3pt]
  \textsuperscript{1}VCIP \& TMCC \& DISSec, College of Computer Science \& \\
  College of Cryptology and Cyber Science, Nankai University
  \\
  \textsuperscript{2}Zhongguancun Academy
  \\
  \textsuperscript{3}Shandong University
  \\
  \texttt{s-yql25@bza.edu.cn},
  \texttt{mengliu.sdu@gmail.com},
  \texttt{yzhou@nankai.edu.cn}
}

\begin{document}

\maketitle
\begingroup
\renewcommand{\thefootnote}{\fnsymbol{footnote}}
\footnotetext[1]{%
Corresponding authors.
}
\endgroup

\begin{abstract}

We explore whether Omni-Language Models (OLMs) can be directly applied to zero-shot Semantic Audio-Visual Navigation (SAVN). Recent work demonstrates that even state-of-the-art specialized models still struggle to achieve generalist multimodal navigation, despite extensive task-specific training. In this paper, we introduce \textbf{RAO-Nav}, short for \textbf{R}easoning \textbf{A}ll-in-One \textbf{O}LM, a deployment pipeline for zero-shot SAVN. By leveraging the rich implicit audio-visual knowledge encoded in OLMs, the embodied agent is enabled to ``\textit{hear}'', ``\textit{see}'', ``\textit{reason}'', and ``\textit{act}'' in the environment. To further elicit the built-in thinking ability of OLMs, we propose a test-time \textbf{L}atent \textbf{N}avigation \textbf{R}easoning (LNR) module that can be seamlessly integrated into the decoding space. LNR encourages the model to retrieve more target-relevant observations and make effective navigation decisions. Through comprehensive experiments, we show that our framework surpasses existing state-of-the-art baselines on public SAVN benchmarks without using any training data. Moreover, we introduce a new \emph{Global Navigation Instruction} setting to further evaluate the ability of OLMs to serve as embodied navigation agents. Code: \texttt{https://github.com/rikeilong/OmniAV\_Nav}
\end{abstract}


\section{Introduction}
Unlike most embodied navigation tasks that focus on egocentric vision \cite{gupta2017cognitive,wu2019bayesian,zhu2017target}, audio-visual navigation \cite{gan2020look,Chen2020SoundSpaces} enables the agent to navigate toward a target location with audio signals in complex real or virtual environments. Furthermore, Semantic Audio-Visual Navigation (SAVN) \cite{Chen2021SAVi,Chen2022SoundSpaces2} extends this task by requiring an agent to use short-duration sound messages to predict what the target is and where it is. It is more challenging not only because the audio signal can be ambiguous, but also because the target may be blocked by walls or other obstacles and cannot always be seen directly \cite{Savva2019Habitat,Szot2021Habitat2}. To deploy in such cases, an embodied agent must go beyond mere visual perception and combine sparse auditory cues with current visual observations to make effective navigational decisions.

Recent progress in Large Language Models (LLMs)~\cite{llama2} has begun to influence embodied robotics~\cite{ahn2022can,chen20232,deitke2020robothor}. Early efforts~\cite{Yang2024RILA} mainly use LLMs as high-level planners or language-centric modules to support navigation. However, it mainly focuses on language-driven perception and scene understanding via collaboration among multi-LLMs, which limits its practicality in real-world deployment. More recently, the development of Omni-Language Models (OLMs\footnote{In this paper, OLMs refer to models capable of processing image, video, and audio inputs.})~\cite{ye2024CAT,baichuan,qwenomni,videollama2} has extended the world knowledge to allow models to process not only language, but also audio and visual inputs \cite{hpl1,anygpt,onellm}. With carefully crafted prompts, they can be adapted to a wide range of downstream tasks without task-specific training.


\begin{figure*}[!h]
    \centering
    \vspace{-1.8em}
    \includegraphics[width=\textwidth]{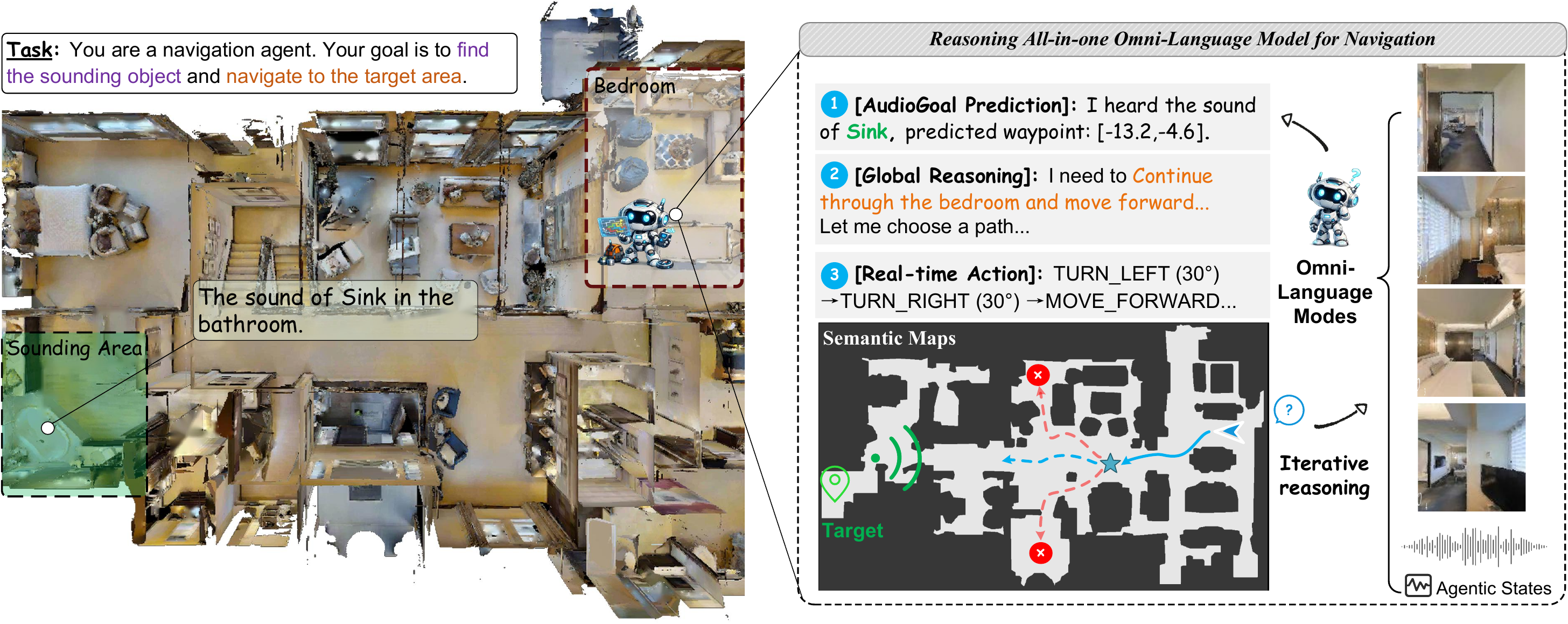}
    \caption{We propose the RAO-Nav pipeline built on a single OLM to achieve zero-shot audio-visual navigation. It can be deployed in both real-world and virtual environments and iteratively performs navigation in three steps: \textit{AudioGoal Prediction}, \textit{Global Reasoning}, and \textit{Real-time Action}. }
    \label{fig:fig1}
\end{figure*}

This naturally raises a question: \textit{can OLMs serve as the main brain for SAVN?} Similar to how humans solve audio-visual navigation by first identifying the sounding object and then moving toward the target, OLMs also require reacting to such tasks with three fundamental aspects: \textbf{1) can the model infer what the sounding target may be and where it is}? \textbf{2) can it form a rough action plan according to the surrounding scene}? \textbf{3) can it act autonomously based on new observations to complete a navigation task}? First, although OLMs can recognize the category of a sound, they do not naturally provide explicit Sound Source Localization (SSL)~\cite{adavanne2018doa,grumiaux2022survey}, making it difficult to ground audio understanding into a precise spatial position. Second, Qwen-Omni is not originally designed for discrete embodied action prediction, so directly mapping multimodal observations to output actions remains unstable. Third, a single OLM decision is often limited by short planning horizons, which makes it prone to spinning in place during action execution. Since OLMs have shown strong multimodal understanding ability, how to effectively elicit such reasoning ability for SAVN remains an open challenge.

To address the aforementioned challenges, we propose \textbf{RAO-Nav}, a deployment pipeline that uses a single OLM to achieve zero-shot SAVN. RAO-Nav decomposes the navigation into an iterative three-stage process: \textit{AudioGoal Prediction}, \textit{Global Reasoning}, and \textit{Real-time Action}. First, OLMs infer the target category via their built-in audio understanding ability. To locate the target, we design a distinct location module to guide the frozen OLM with dual-frequency spectrograms. Then, the model performs a high-level navigation guidance (e.g., leave the current room and follow the hallway for a long stretch) with the semantic map. At the execution step, the model receives continuous RGB observations and updates the action sequence in the environment. In this way, RAO-Nav allows a single OLM to connect multimodal perception, planning, and action within one unified pipeline. To further improve the reasoning ability of the OLM during navigation, we introduce a test-time \textbf{L}atent \textbf{N}avigation \textbf{R}easoning (LNR) module. LNR's goal is to encourage the model to focus on more target-relevant observations during global reasoning. As a result, the model can make more stable and effective decisions in long-horizon and uncertain environments.

We compare our RAO-Nav with existing specialized models and multi-LLM collaboration frameworks on public SAVN benchmarks. Results show that our method outperforms most existing baselines without using any task-specific training data. Moreover, the single-OLM paradigm offers substantial convenience for real-world deployment, as an embodied agent can be driven to complete navigation tasks using only a server equipped with a single GPU. To further test whether OLMs can serve as practical embodied agents beyond simulation platform settings, we introduce a new Global Navigation Instruction setting based on SAVN. It evaluates the model under more realistic language-guided global perception and navigation scenarios. The results show that, although OLMs demonstrate promising zero-shot ability, deploying them in real-world natural-language-guided audio-visual navigation remains challenging. Our contributions are summarized as follows:
\begin{itemize}
    \item We propose RAO-Nav for SAVN, exploiting the powerful audio-visual understanding ability of OLMs to navigate effectively.
    \item Building on this pipeline, we further introduce a test-time LNR module to better elicit the reasoning ability of OLMs during navigation.
    \item Experiments on public SAVN benchmarks show that our method achieves clear zero-shot improvements over strong baselines without using any task-specific training data.
\end{itemize}

\section{Related Work}
\noindent \textbf{Semantic Audio Visual Navigation.} SAVN has developed with the emergence of embodied simulators and realistic audio-visual environments (such as Habitat~\cite{Savva2019Habitat,Szot2021Habitat2} and SoundSpaces~\cite{Chen2020SoundSpaces,Chen2022SoundSpaces2}). Early research~\cite{Chen2021SAVi} primarily focused on learning end-to-end navigation strategies by jointly modelling visual and auditory observations. For example, SAVi~\cite{Chen2021SAVi} introduces a benchmark for evaluating whether a model can drive navigation to audible semantic targets. These works establish the basic problem setting, where the agent must infer the target from sparse audio cues and navigate under partial visual observations. Recent research has further advanced this field in several directions, including modular transfer learning~\cite{AlHalah2022ZeroExperience}, knowledge-enhanced scene priors~\cite{Tatiya2022KnowledgeDriven}, and more realistic scenarios incorporating moving sound sources in complex, unmapped environments~\cite{Younes2023CatchMe}. Such extensions push SAVN beyond static indoor settings and make the task closer to real-world deployment. In addition to standard SAVN, AVLEN~\cite{Paul2022AVLEN} extends the task to audio-visual-language embodied navigation, exploring oracle-language to support navigation. These efforts show that SAVN is gradually moving from task-specific policy learning toward richer multimodal reasoning and interaction.


\noindent \textbf{SAVN with MLLMs.} MLLMs~\cite{ye2024CAT,ye2025eyes,ye2025cat+,ye2024pose,ye2025sugar} refer to models that understand both video and audio inputs and output text instructions. Recent studies~\cite{Yang2024RILA,zhou2023esc} investigate whether language-centric multimodal agents can perform navigation through prompt engineering. RILA~\cite{Yang2024RILA} introduces a reflective and imaginative language agent for zero-shot SAVN, showing that LLM-based planning can leverage multimodal perception to guide exploration. This line of work suggests that large models may provide a new alternative to heavily trained navigation policies. NavBench~\cite{Qiao2025NavBench} probes the navigation capability of MLLMs under zero-shot settings, and the results show that although open-source MLLMs exhibit promising spatial and action reasoning ability, they still struggle with robust long-horizon decision-making. Moreover, OctoNav~\cite{Gao2025OctoNav} explores free-form embodied navigation and argues that it is difficult for MLLMs to achieve generalist embodied navigation. These findings indicate that strong multimodal understanding alone does not directly translate into reliable embodied control. Therefore, we believe that applying MLLMs designed for general-purpose scenarios to SAVN remains a challenge that has not yet been fully explored.

\section{RAO-Nav Pipeline}

\begin{figure*}[!h]
    \centering
    \includegraphics[width=\textwidth]{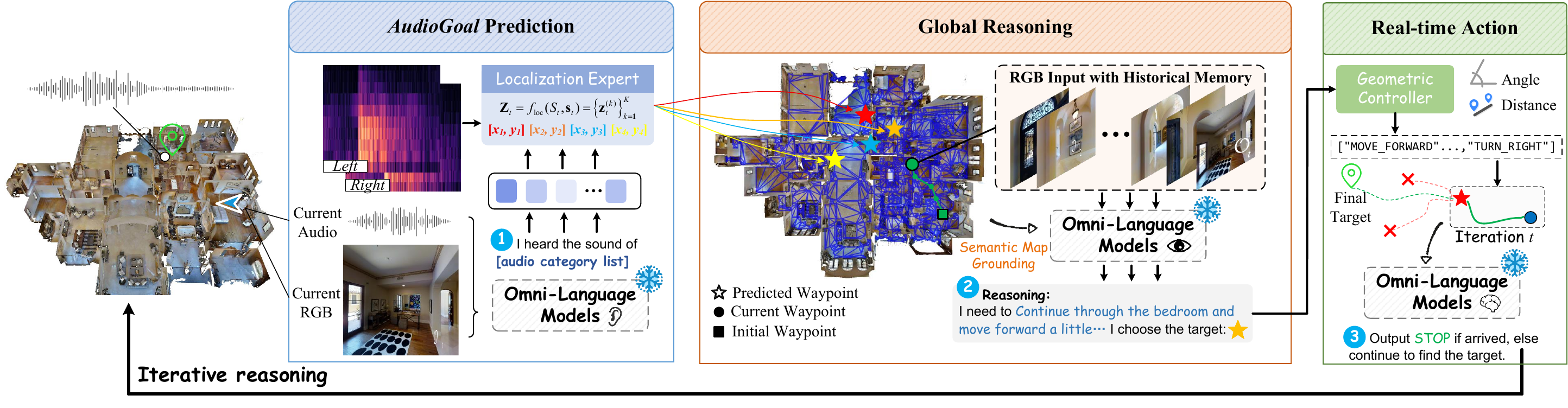}
    \caption{\textbf{Overview of the RAO-Nav Pipeline for zero-shot SAVN.} Given a scene and a queried episode, the agent receives the current RGB observation and the incoming audio signal (or \texttt{None} if no new audio is observed). \includegraphics[scale=0.38]{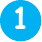} The OLM first performs audio classification and predicts the target waypoint at time step $t$ with the Localization Expert. Conditioned on multiple target waypoints and historical RGB observations, \includegraphics[scale=0.38]{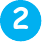} the OLM performs language-driven navigation reasoning to choose a target. The selected target is executed by a low-level action compiler, \includegraphics[scale=0.38]{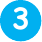} while the OLM continuously determines whether the agent should stop based on the current surroundings. Notably, all OLMs in the pipeline share the same model.}
    \label{fig:framework}
\end{figure*}

In this section, we study whether a single OLM can solve the SAVN task~\cite{Chen2021SAVi} in a zero-shot manner. As shown in Fig. \ref{fig:framework}, the agent is required to identify a sounding target and navigate toward it in complex and unseen environments by jointly using auditory and visual observations. Instead of relying on task-specific training trajectories, we aim to directly leverage the intrinsic multimodal perception and reasoning ability of OLMs for navigation. To this end, we propose \textbf{RAO-Nav}, a unified deployment pipeline that enables the model to perform \textit{AudioGoal Prediction}, \textit{Global Reasoning}, and \textit{Real-time Action} in a human-like step-by-step manner.

\subsection{Problem Definition}
Consider an agent in a previously unseen 3D indoor environment, where navigation is performed over a finite action space $\mathcal{A}=\{\texttt{stop}, \texttt{move\_forward}, \texttt{turn\_left}, \texttt{turn\_right}\}$. The goal of SAVN is to locate and reach a sounding target object in the scene. To better adapt the formulation to OLMs, we do not introduce redundant components used in previous work \cite{Paul2022AVLEN}, such as the immediate reward, the probability of observation, or the transition probability. Instead, we formulate the sequential decision process characterized by the tuple $(\mathcal{H},\mathcal{S},\mathcal{A},\mathcal{O})$. At each time step $t$, $\mathcal{O}_t$ denotes the current ego-centric RGB image, $a_t \in \mathcal{A}$ denotes the audio signal, and $s_t \in \mathcal{S}$ denotes agent states such as angle and distance to the predicted goal. Similar to \cite{Chen2021SAVi}, we keep a history of past observation $\mathcal{H}_t=\{\mathcal{O}_1,\ldots,\mathcal{O}_t\}$ as the memory for reasoning. Based on these inputs, OLMs as agents continuously update the current \textit{AudioGoal} to maximize the probability of reaching the sounding target and stopping at the correct location.

\subsection{\textit{AudioGoal} Prediction}

Fully open-set audio classification for OLMs remains challenging, as reverberation and semantically similar acoustic events are common in both virtual and real-world environments. Since an audio signal may only be present momentarily, we propose an iterative perception–localization strategy built upon OLMs, enabling them to reason about the target object in a more generalizable way for SAVN. Specifically, it is divided into two steps:

\noindent \textbf{Perception.} Let $\mathcal{A}_t$ denote the audio observation received at time step $t$, and let $\mathbf{u}_t$ denote the action executed at this step. We use
$\mathcal{A}_{1:i}=\{\mathcal{A}_1,\ldots,\mathcal{A}_i\}$ and $\mathbf{u}_{1:i}=\{\mathbf{u}_1,\ldots,\mathbf{u}_i\}$ to represent the audio observations and action history up to step $i$, respectively. Given the accumulated audio observations and action history, the OLM
produces a $\mathcal{G}$-dimensional vector of audio-category scores:
\begin{equation}
\mathbf{s}_i =
G_{\mathrm{obj}}\left(\mathcal{A}_{1:i},\mathbf{u}_{1:i}\right)
\in \mathbb{R}^{|\mathcal{G}|}
\end{equation}
where $\mathcal{G}$ is the fixed set of the defined audio categories in the SAVi benchmark, and $s_{i,g}$ denotes the score assigned to category $g\in\mathcal{G}$ at step $i$. These scores are generated directly by the frozen OLM without task-specific training or fine-tuning. 


Since an audio signal may only be available for a short period, we model audio prediction as a progressively updated belief. After each executed action, the OLM updates the category scores using the newly accumulated audio observations. We then construct the candidate target set as

\begin{equation}
\hat{\mathcal{G}}_t =
\left\{
g\in\mathcal{G}\;\middle|\;
\sum_{i=1}^{t}
\max\left(0,\;s_{i,g}-\tau\right)
\geq \gamma
\right\},
\end{equation}

where the summation ranges from the first step to the current step ($i=1,\ldots,t$). In all experiments, we set $\tau=0.12$ and $\gamma=0.5$. Since the uniform score is $1/21\approx 0.048$, $\tau$ filters out weak per-step predictions, whereas $\gamma$ retains categories with sufficiently strong accumulated evidence. The resulting candidate set $\hat{\mathcal{G}}_t$ is then used by the localization module.

\noindent \textbf{Localization.} General-purpose OLMs can recognize the semantic category of a sound but do not reliably estimate its spatial location. To provide explicit spatial grounding, we use a plug-in Localization Expert following the localization design of~\cite{Chen2021SAVi}. The expert is used without additional task-specific training and predicts the relative source position from binaural audio features and the OLM-generated category information.

Let $\mathcal{A}_{t,L}$ and $\mathcal{A}_{t,R}$ denote the left and right audio channels at time step $t$. We first convert each channel into a time-frequency representation and apply logarithmic compression:
\begin{equation}
S_{t,c}=\log\left(1+F(\mathcal{A}_{t,c})\right),
\qquad c\in\{L,R\},
\end{equation}
where $F(\cdot)$ denotes the time-frequency transformation. We then stack the two spectrograms $
S_t=\mathrm{Stack}(S_{t,L},S_{t,R}).$ This two-channel representation preserves both spectral information and inter-channel differences.

To condition localization on the OLM's semantic prediction, we use the continuous $\mathcal{G}$-dimensional category-score vector $\mathbf{s}_t$ as the localization input. Then, the Localization Expert predicts $K$ possible source positions:
\begin{equation}
\mathbf{Z}_t
=
f_{\mathrm{loc}}(S_t,\mathbf{s}_t)
=
\left\{\mathbf{z}_t^{(k)}\right\}_{k=1}^{K},
\qquad
\mathbf{z}_t^{(k)}
=
\left[u_t^{(k)},v_t^{(k)}\right].
\end{equation}

Each predicted position is then transformed into the agent coordinate system:$
\hat{\mathbf{c}}_t^{(k)}
=
\begin{bmatrix}
-v_t^{(k)}\\
u_t^{(k)}
\end{bmatrix},$ where $\hat{\mathbf{c}}_t^{(k)}$ is the relative position of the
$k$-th possible sound source. In this way, the expert only provides geometric estimation conditioned on the predicted target for navigation, allowing our RAO-Nav pipeline to retain the audio understanding ability of OLMs.




\subsection{Global Reasoning}
Although OLMs hold powerful transferability, they are still fundamentally language-driven models and are not well-suited for directly predicting discrete actions. To make the planning more robust while keeping the decision space tractable for OLMs, we follow \cite{Gao2025OctoNav} and introduce a QA-driven reasoning paradigm with semantic map grounding. Specifically, given the candidate waypoint set $\mathbf{Z}_t$ at iteration $t$ and the current agent position $\mathbf{w}_t^{\text{pos}}$, the OLM needs to reason jointly over the predicted audio object and the visual observations to determine which path is most consistent with the target semantics, instead of reasoning over the entire discrete action space. We present simplified versions of the task-specific prompts below.

\begin{figure}[!h]
	\centering
    \vspace{-0.5em}
	\includegraphics[scale=0.718]{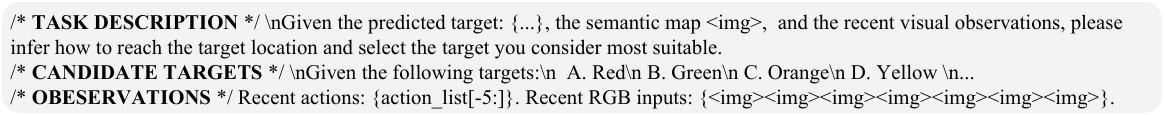}
    \vspace{-1.8em}
\end{figure}

This explicit multi-waypoints formulation makes better use of the zero-shot decision-making ability of OLMs, while encouraging the agent to explore sufficiently diverse routes for completing SAVN.

\subsection{Real-time Action}
To convert the selected path into executable low-level actions, we introduce a \textit{Geometric Controller} that compiles the trajectory into a deterministic action sequence. Instead of predicting one or a few actions from local observations at each step, we execute long-horizon actions with the reasoning of OLMs. Given the selected path $P_t^{(z_t)}=\{\mathbf{w}_i^{\text{pos}}\}_{i=0}^{N_t}$, where $\mathbf{w}_0^{\text{pos}}$ is the current position and $\mathbf{w}_{N_t}^{\text{pos}}$ is the predicted goal, we decompose each $\mathbf{w}_0^{\text{pos}} \rightarrow \mathbf{w}_{N_t}^{\text{pos}}$ into a \emph{turn-then-forward} procedure. Specifically, we compute the turning angle $\theta_i$ between the current heading and the segment direction, and the travel distance $d_i$ on the $xz$ plane, which can be denoted as:
\begin{equation}
n_i^{\text{turn}} = \left\lfloor \frac{|\theta_i|}{\Delta\theta} \right\rceil, 
\quad
n_i^{\text{forward}} = \left\lfloor \frac{d_i}{\Delta s} \right\rceil,
\end{equation}
where $\Delta\theta$ and $\Delta s$ denote the fixed rotation angle and forward step size.

After executing each compiled action sequence, the agent moves to a new position and receives updated RGB and audio observations. Furthermore, to stabilize localization across iterations, we apply temporal smoothing to every predicted point goal:
\begin{equation}
\tilde{\mathbf{z}}_t=(1-\omega)\mathbf{z}_t+\omega\,\Pi(\tilde{\mathbf{z}}_{t-1}),
\qquad \omega\in[0,1],
\end{equation}
where $\Pi(\cdot)$ projects the previous estimate into the current coordinate frame. When no new audio is observed, we retain the previous estimate instead of forcing a new prediction. During execution, the OLM continuously monitors the environment and decides whether to terminate with \texttt{STOP}. Otherwise, the system re-runs \textit{AudioGoal} prediction and generates a new candidate path, forming a closed-loop predict--plan--decide--execute pipeline.



\section{Test-time Latent Navigation Reasoning Module}

To reduce the amount of historical RGB observations used in subsequent iterations, we introduce a simple yet effective test-time LNR module that enables the OLM to selectively focus on target-relevant observations without additional training. Specifically, the agent accumulates a sequence of historical RGB embeddings $\mathcal{V} = \texttt{get\_input\_embeddings}( \mathcal{O}_1,\ldots,\mathcal{O}_t)$. \begin{wrapfigure}[12]{r}{0.55\textwidth}
  \vspace{-1.0em}
  \centering
  \includegraphics[width=0.55\textwidth]{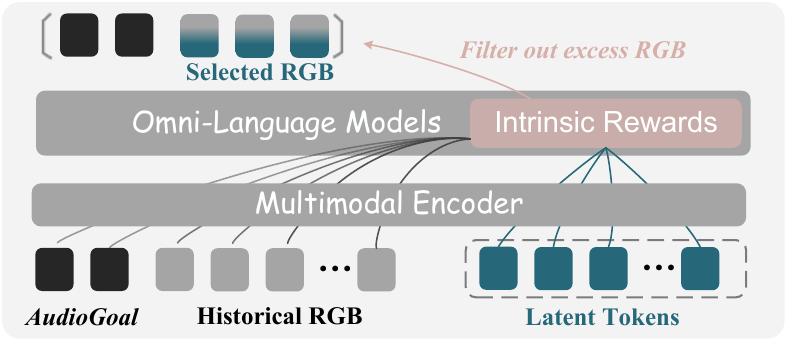}
  \vspace{-1.8em}
  \caption{Overview of the proposed LNR module.}
  \label{fig:data_prepare}
\end{wrapfigure}Then, we initialize $L$ latent tokens on the text side, each represented by a special placeholder token \texttt{<|latent|>}. These tokens are embedded using the same \texttt{get\_input\_embeddings()} function and concatenated with other multimodal tokens. Formally, these embeddings are denoted as $\mathbf{H} = [h_1, h_2, \dots, h_L]$ and share the same embedding space as other tokens. Instead of updating model parameters, we define an intrinsic reward based on the model’s conditional likelihood:
\begin{equation}
R(\mathbf{H}^{(t)}) = \log p_\theta(\hat{\mathcal{G}}_t^k \mid \mathcal{V}, \mathbf{H}^{(t)}),
\end{equation}
where $\hat{\mathcal{G}}_t^k$ denotes the predicted audio goal (represented as text tokens). A higher likelihood indicates that the latent tokens better summarize goal-relevant information from the observations. Then, we iteratively update the latent tokens to maximize this reward by:
\begin{equation}
\mathbf{H}^{(t+1)} = \mathbf{H}^{(t)} + \eta \nabla_{\mathbf{H}} R(\mathbf{H}^{(t)}).
\end{equation}
where the gradient $\nabla_{\mathbf{H}}$ is computed via backpropagation through the frozen OLM, and $\eta$ is a step size. After optimization, the latent tokens are aggregated into a query $q = \text{MeanPool}(\mathbf{H}).$ Finally, we compute the similarity between the query and each frame embedding to select the top-$K$ RGB embeddings to form a compact subset for the next navigation reasoning step.

\section{Experiments}
\subsection{Experiment Setup}
\noindent \textbf{Simulator and Datasets.} We evaluate our method in the SoundSpaces~\cite{Chen2020SoundSpaces} simulator built on Habitat~\cite{Savva2019Habitat}. The 3D environments are constructed from the Matterport3D (MP3D) dataset \cite{chang2017matterport3d}, which provides realistic indoor layouts and object labels. Unlike prior SAVN works \cite{Chen2021SAVi,Paul2022AVLEN}, we focus on unheard sounds and unseen scenes at test time, where the agent must navigate toward sounding target objects without access to task-specific training trajectories. This setting is suitable for evaluating whether a single OLM can generalize to navigation tasks in a zero-shot manner.

\noindent \textbf{Baselines.} We evaluate four different types of baselines, including:
 \vspace{-0.5em}
\begin{itemize}[leftmargin=0.5em]
    \item \textbf{Specialized Models} are task-specific navigation methods trained with environment trajectories, including AudioGoal~\cite{Chen2020SoundSpaces}, SAVi~\cite{Chen2021SAVi}, AV-WAN~\cite{chen2021waypoint}, AVLEN~\cite{Paul2022AVLEN}, and K-SAVEN~\cite{Szot2021Habitat2}. These methods are usually optimized with task-specific supervision and rely on large amounts of environment interaction data.
    \vspace{-0.3em}
    \item \textbf{Multi-MLLMs Collaborative Frameworks} perform zero-shot navigation through the cooperation of multiple language-based modules, e.g., RILA~\cite{Yang2024RILA}. In this paradigm, different modules are designed for separate roles, such as audio-visual perception and global planning. While this design can achieve zero-shot SAVN, it increases system complexity and makes real-world deployment less convenient.
    \vspace{-0.3em}
    \item \textbf{Pure Vision-Language Models (VLMs) with Audio Descriptions.} This setup is designed to assess whether VLMs possess visual navigation capabilities after a given fixed audio target. We feed audio information as textual descriptions rather than directly reasoning over raw audio-visual inputs. The evaluation model includes Qwen-VL Series (7B/8B/30B)~\cite{qwenvl,bai2025qwen3vl}, and LLaVA-NeXT~\cite{liu2023llava}. 
    \vspace{-0.3em}
    \item \textbf{OLMs w/ RAO-Nav Pipeline} use one unified model to jointly process audio, vision, and interaction history for navigation. We evaluate models including Qwen2.5-Omni-7B~\cite{qwenomni}, and Baichuan-Omni-1.5-7B~\cite{baichuan}.
\end{itemize}
 \vspace{-0.5em}
   Through this comparison, we examine how RAO-Nav differs from heavily trained specialized models as well as from different zero-shot paradigms.

\noindent \textbf{Metrics.} We follow standard SAVN evaluation protocols and report Success Rate (SR), Success weighted by Path Length (SPL), Success Rate weighted by Number of Actions (SNA), Success When Silent (SWS), and Distance To Goal (DTG). Moreover, we further evaluate the performance of OLMs in the global navigation instruction setting in Section 5.3.

\noindent \textbf{Implementation Details.} Consistent with the standard SAVN setting, the agent is provided with RGB observations at a resolution of $256 \times 256$. It also receives audio signals in \texttt{wav} format, which are further converted into two-channel spectrograms of size $65 \times 26$ for goal prediction. All large models in our framework are deployed on only one NVIDIA A100 GPU (80GB). We evaluate RAO-Nav on 1,000 test episodes, and limit the maximum number of execution steps to 200 for each episode.





\begin{table*}[t]
\small
\centering
\caption{Comparison of four types of navigation frameworks on the Matterport3D test dataset in unheard-sound settings. Oracle$^{\dagger}$ denotes a privileged-input diagnostic setting for Pure VLMs. The model is given the ground-truth audio category, the semantic map, and the ground-truth target location marked on the map. It directly predicts low-level actions.}
\label{tab:main_unheard}
\setlength{\tabcolsep}{0.5mm}
\begin{tabular}{llccccc}
\toprule
& Method & SR (\%)$\uparrow$ & SPL (\%)$\uparrow$ & SNA (\%)$\uparrow$ & DTG (m)$\downarrow$ & SWS (\%)$\uparrow$ \\
\rowcolor{gray!10}\multicolumn{7}{c}{\textit{\textbf{Supervised}}} \\
\multirow{5}{*}{Specialized Models}
& AudioGoal    & 16.5 & 15.5 & 10.4 & 12.8 & 5.6 \\
& AV-WAN       & 17.2 & 13.2 & 12.7 & 11.0 & 6.9 \\
& SAVi         & 24.8 & 17.2 & 13.2 & 9.9  & 14.7 \\
& AVLEN       & 26.2 & 17.6 & 14.2 & 9.2  & 15.8 \\
& K-SAVEN     & 34.4 & \textbf{23.4} & \textbf{21.7} & \textbf{6.6} & 14.3 \\
\rowcolor{gray!10}\multicolumn{7}{c}{\textit{\textbf{Zero-Shot}}} \\
Multi-MLLMs& RILA & 35.4  & 11.8  & 8.7 & 11.4 & 20.4 \\
\midrule
\multirow{4}{*}{Pure VLMs}& LLaVA-NeXT + Oracle$^{\dagger}$ & 4.6 & 2.9  & 2.4 & 19.6 & 2.1 \\
& Qwen2.5-VL-7B + Oracle$^{\dagger}$& 8.6  & 4.8  & 3.7 & 28.7 & 4.3 \\
& Qwen3-VL-8B + Oracle$^{\dagger}$& 7.9  & 2.5  & 4.2 & 21.3 & 3.8 \\
& Qwen3-VL-30B + Oracle$^{\dagger}$ & 14.7 & 5.2  & 7.6 & 17.6 & 3.5 \\
\midrule
\multirow{2}{*}{\textbf{w/ RAO-Nav}}& Baichuan-Omni-7B& 37.6 & 16.7  & 9.4 & 15.2 & 21.2 \\
& Qwen2.5-Omni-7B& 36.4 & 18.4  & 10.2 & 11.7 & 21.4 \\
\midrule
\multirow{2}{*}{\textbf{w/ RAO-Nav + LNR}}& Baichuan-Omni-7B& 38.4 & 17.5  & 9.6 & 13.7 & \textbf{22.4} \\
& Qwen2.5-Omni-7B& \textbf{39.2} & 20.7  & 12.4 & 10.9 & 21.9 \\

\bottomrule
\end{tabular}
\end{table*}

\subsection{Experimental Results}

\noindent \textbf{Main Experiment.} We compare different navigation frameworks on the Matterport3D dataset under the unheard-sound setting in Table~\ref{tab:main_unheard}. For supervised methods, K-SAVEN achieves the best SPL and SNA by leveraging oracle guidance, indicating that task-specific training helps the agent follow near-optimal trajectories. However, such approaches rely heavily on curated data and generalize poorly to unseen environments. In the zero-shot setting, RILA achieves a higher SR than all supervised baselines, but suffers from low SPL and SNA, suggesting inefficient and unstable navigation. Moreover, directly applying VLMs performs significantly worse, with SR below 15\%, indicating that mapping visual observations to actions without structured reasoning is insufficient. 

In contrast, our RAO-Nav pipeline enables a single OLM to perform SAVN effectively. Without LNR, RAO-Nav already achieves strong performance (e.g., 37.6\% SR with Baichuan-Omni-7B). With LNR, Qwen2.5-Omni-7B further improves to the best SR of 39.2\%, surpassing all supervised baselines, while maintaining competitive SPL and strong stopping performance SWS. This shows that LNR helps filter redundant RGB observations and improves decision stability. However, zero-shot methods still lag behind supervised models in SPL, SWS, and DTG, indicating less efficient trajectories and occasional redundant actions.

\begin{table*}[t]
\small
\centering
\caption{Experimental results on SAVN in our Global Navigation Instruction settings. Closed and Open indicate whether candidate audio categories are provided, and Map indicates whether a semantic map is provided.}
\label{tab:global}
\setlength{\tabcolsep}{2.5mm}
\begin{tabular}{lccccc}
\toprule
\multirow{2}{*}{Methods} &
\multicolumn{2}{c}{\textit{Audio Classification}} &
\multicolumn{2}{c}{\textit{Visual Navigation}} &
\multirow{2}{*}{All avg.} \\
\cmidrule(lr){2-3}\cmidrule(lr){4-5}
& Closed & Open 
& w/o Map & w/ Map
& \\
\midrule
\rowcolor{gray!10}\multicolumn{6}{c}{\color{darkedred}{\textit{OLMs with an average accuracy below 40\% $\downarrow$}}} \\

MIO-Instruct-7B $_{\color{gray}\textit{(open-source)}}$&
47.56 & 6.40 &
18.95 & 17.74 & 22.84 \\

PandaGPT-7B $_{\color{gray}\textit{(open-source)}}$&
45.73 & 9.88 &
25.81 & 28.63 & 27.04 \\

Video-SALMONN-o1-7B $_{\color{gray}\textit{(open-source)}}$&
53.05 & 12.21 &
29.03 & 30.65 & 31.25 \\

Video-LLaMA2-7B $_{\color{gray}\textit{(open-source)}}$&
40.85 & 5.23 &
23.79 & 27.02 & 24.28 \\

\rowcolor{gray!10}\multicolumn{6}{c}{\color{Forestgreen}{\textit{OLMs with an average accuracy above 40\% $\uparrow$}}} \\

Baichuan-Omni-7B $_{\color{gray}\textit{(open-source)}}$&
59.76 & 23.84 &
35.89 & 45.56 & 40.99 \\

Qwen2.5-Omni-7B $_{\color{gray}\textit{(open-source)}}$&
65.24 & 21.51 &
37.10 & 43.15 & 42.67 \\

ChatGPT-4o $_{\color{gray}\textit{(closed-source)}}$&
75.61 & 36.05 &
41.94 & 53.23 & 48.80 \\

Gemini2.0-Flash $_{\color{gray}\textit{(closed-source)}}$&
71.34 & 31.40 &
50.81 & 52.02 & 51.68 \\

\bottomrule
\end{tabular}
\end{table*}

\noindent \textbf{Experiment on Global Navigation Instruction.} To assess whether current OLMs are suitable for our proposed RAO-Nav deployment, we evaluate their performance under a Global Navigation Instruction setting within the SAVN task. We collect 164 and 172 samples for the closed-set and open-set in the audio classification task, respectively, and 248 samples for the visual navigation task. This setting highlights language-driven navigation, where the agent must rely on high-level instructions rather than step-by-step supervision. We first examine the audio perception capability of OLMs, which is essential for grounding the target sound source. As shown in Table \ref{tab:global}, we evaluate models under both closed-set and open-set conditions. In the closed setting, candidate audio categories are provided, while in the open setting, models must infer the target category without prior constraints. The results show that although several OLMs achieve reasonable performance in the closed setting, there is a clear performance drop in the open setting, indicating that robust audio understanding and generalization remain challenging for most models.

Furthermore, we investigate how well OLMs align visual observations with language instructions. Each input consists of multiple candidate navigation options, each paired with a corresponding RGB observation. The model is required to select the option that best matches the target. This setup evaluates whether the model can jointly reason over visual content and language descriptions. The results indicate that most models struggle to consistently identify the correct option, especially in complex scenes with subtle visual differences. Even strong models tend to be distracted by irrelevant visual cues or fail to correctly associate language instructions with the corresponding observations. This suggests that cross-modal alignment between vision and language instruction remains a key bottleneck for OLMs in navigation tasks.

\subsection{Ablation Studies}
\noindent \textbf{Ablation on Localization Expert.}
We compare three localization strategies to explore the impact of target position estimation. \textit{Random} refers to sampling target locations randomly on the map, while \textit{OLM-only Localization} lets the model infer the target position directly from the predicted audio category and the semantic map. As shown in Table~\ref{tab:ablation}(a), the result shows that current general-purpose OLMs are weak at sound-source localization, which motivates the use of the Localization Expert. We study whether audio-category information is useful for localization. Following the two SAVi variants \cite{Chen2021SAVi}, we compare a spectrogram-only localizer with a localizer that receives the OLM-predicted audio-category information. The result shows that removing the category input reduces the SR from 39.2\% to 12.6\%. We believe that reliable localization benefits from both acoustic and semantic information. 

\noindent \textbf{Ablation on Input Modalities.}
We further investigate the effect of input modalities in Table~\ref{tab:ablation}(b). Interestingly, using only RGB observations achieves the best performance, while incorporating depth information results in a drop in SR and SPL and an increase in DTG. We argue that depth images do not provide additional useful cues for the OLM. Instead, it introduces extra input complexity and increases the reasoning burden of the model. Since RAO-Nav relies on language-driven reasoning over multimodal inputs, adding depth may distract the model from focusing on more informative semantic and visual signals. Therefore, we adopt a simplified design that only uses RGB inputs, which is more effective for zero-shot navigation.

\noindent \textbf{Ablation on LNR.}
We compare LNR with two history-selection baselines: randomly selecting $K$ observations (\textit{Random-$K$}) and selecting the $K$ nearest observations in the embedding space (\textit{$K$-NN}). Compared with $K$-NN, LNR improves SR and SPL by 1.4\% and reduces DTG by 1.5 m, showing that target-guided latent optimization is more effective than simple observation selection.




\subsection{Visualization of Navigation Trajectories}
\vspace{-0.8em}

\begin{figure*}[t]
    \centering
    \includegraphics[width=\textwidth]{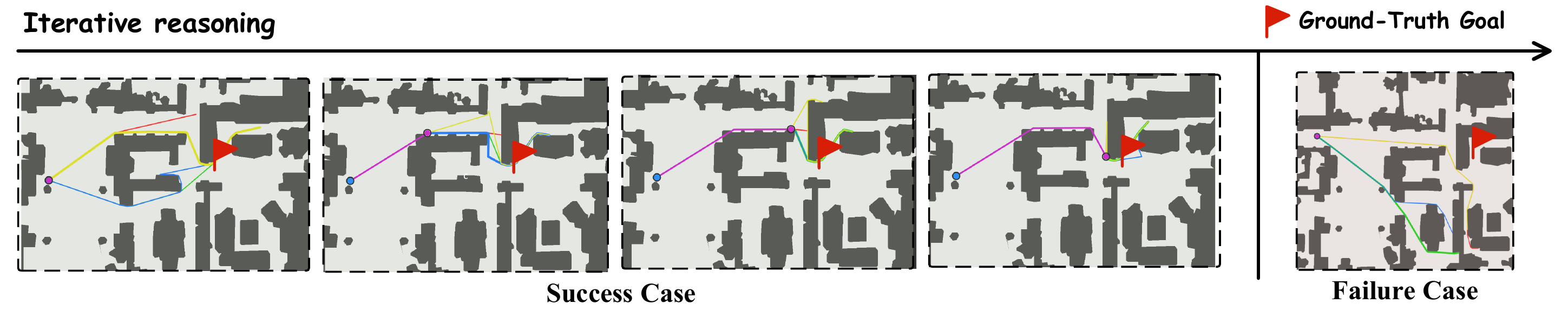}
    \caption{Navigation trajectories in different iterations with RAO-Nav.}
    \label{fig:visual}
\end{figure*}

To better understand the behavior of RAO-Nav, we visualize navigation trajectories under the iterative reasoning process, including both success and failure cases. As shown in Fig.~\ref{fig:visual}, in successful cases, the agent is able to progressively refine its decisions by continuously incorporating new visual observations. Although the initial trajectories may deviate from the ground-truth goal due to coarse localization, the model gradually corrects its direction through iterative reasoning and ultimately reaches the sounding target. In contrast, most failure cases are primarily caused by inaccurate sound source localization. When the predicted target position is incorrect, the subsequent reasoning process is guided toward a wrong region, leading the agent to follow suboptimal or misleading paths. As a result, even though the model continues to update its reasoning, it fails to recover and reach the true target. It suggests that while OLMs have strong potential for navigation through iterative reasoning, their performance heavily depends on the accuracy of target localization, highlighting the importance of reliable spatial grounding in SAVN.



\begin{table*}[t]
\centering
\caption{\textbf{Ablations.} We conduct detailed ablations based on Qwen2.5-Omni-7B in SAVN.}
\label{tab:ablation}
\setlength{\tabcolsep}{5pt}
\begin{minipage}[t]{0.49\textwidth}
\centering
\small
\caption*{(a) Results of different localization strategies. All settings use the same OLM for global reasoning and action decisions. $\hat{\mathcal{G}}_t$ denotes the OLM reasoning output.}
\vspace{-0.8em}
\begin{tabular}{lccc}
\toprule
\makecell{Method} & SR & SPL & DTG \\
\midrule
Random & 3.4 & 1.9 & 21.7 \\
OLM-only Localization & 10.7 & 5.8 & 16.4\\
Localization Expert (w/o $\hat{\mathcal{G}}_t$) & 12.6 & 7.4 & 14.6 \\
Localization Expert (w/ $\hat{\mathcal{G}}_t$) & \textbf{39.2} & \textbf{20.7} & \textbf{10.9} \\
\bottomrule
\end{tabular}
\end{minipage}
\hfill
\begin{minipage}[t]{0.49\textwidth}
\centering
\small
\caption*{(b) Results of different input modalities.}
\vspace{-0.8em}
\begin{tabular}{lccc}
\toprule
\makecell{Modality} & SR & SPL & DTG \\
\midrule
only RGB  & \textbf{39.2} & \textbf{20.7} & \textbf{10.9} \\
w/ Depth & 28.6 & 12.4 & 18.4 \\
\bottomrule
\end{tabular}

\vspace{0.2em}

\caption*{(c) Results of different LNR variants.}
\vspace{-0.8em}
\setlength{\tabcolsep}{1.2mm}{
\begin{tabular}{lccc}
\toprule
\makecell{Method} & SR& SPL & DTG  \\
\midrule
Random-$K$ & 34.6 & 18.3  & 13.7  \\
$K$-NN & 37.8 & 19.3  & 12.4 \\
LNR & \textbf{39.2} & \textbf{20.7} & \textbf{10.9} \\
\bottomrule
\end{tabular}%
}
\end{minipage}

\end{table*}

\section{Conclusion}
\vspace{-0.8em}

In this paper, we investigate whether OLMs can serve as a unified solution for zero-shot SAVN. We propose RAO-Nav, a simple yet effective pipeline that enables a single OLM to jointly perform audio understanding, global reasoning, and action execution in an iterative manner. By decomposing navigation into \textit{AudioGoal} Prediction, Global Reasoning, and Real-time Action, our framework bridges multimodal perception and decision-making without any task-specific training. Furthermore, we introduce a test-time LNR module to filter redundant historical RGB observations and encourage the model to focus on target-relevant information during iterative reasoning. Experimental results on SAVN benchmarks demonstrate that RAO-Nav achieves strong zero-shot performance and surpasses existing supervised baselines.

\section{Acknowledgments}
This research was funded by the Zhongguancun Academy (Grant No. 20240306), the National Natural Science Foundation of China (Grant NO 62376266 and 62406318), CAAI-Tencent Rhino-Bird Open Research Fund, and CAAI-Ant Group Research Fund.



{\small
\bibliographystyle{neurips_like_refs}
\bibliography{papers}
}


\appendix



\end{document}